\documentclass[journal,comsoc]{IEEEtran}
\usepackage[T1]{fontenc}
\usepackage{cite}

\ifCLASSINFOpdf
   \usepackage[pdftex]{graphicx}
\else
   \usepackage[dvips]{graphicx}
\fi

\usepackage{amsmath}
\usepackage[cmintegrals]{newtxmath}

\usepackage[numbers]{natbib}
\usepackage{booktabs}
\usepackage{multirow}
\usepackage{multicol}
\usepackage{algorithm}
\usepackage{algorithmic}

\usepackage{mathtools}

\usepackage{enumitem}

\mathchardef\mhyphen="2D

\begin{document}
%
\title{Scatter-Based Innovation Propagation in Large Language Models for Multi-Stage Process Adaptation}

\author{Hong~Su
\IEEEcompsocitemizethanks{\IEEEcompsocthanksitem H. Su is with the School of Computer Science, Chengdu University of Information Technology, Chengdu, China.\\
 E-mail: suguest@126.com. \\
\protect\\
}
\thanks{}}

\markboth{Journal of \LaTeX\ Class Files,~Vol.~14, No.~8, August~2015}%
{Shell \MakeLowercase{\textit{et al.}}: Bare Demo of IEEEtran.cls for IEEE Communications Society Journals}
%

\maketitle

\begin{abstract}
Large Language Models (LLMs) exhibit strong capabilities in reproducing and extending patterns observed during pretraining but often struggle to generalize novel ideas beyond their original context. This paper addresses the challenge of applying such localized innovations—introduced at a specific stage or component—to other parts of a multi-stage process. We propose a scatter-based innovation expansion model (innovation scatter model) that guides the LLM through a four-step process: (1) identifying the core innovation by comparing the user’s input with its surrounding context, (2) generalizing the innovation by removing references to specific stages or components, (3) determining whether the generalized innovation applies to a broader scope beyond the original stage, and (4) systematically applying it to other structurally similar stages using the LLM. This model leverages structural redundancy across stages to improve the applicability of novel ideas. Verification results demonstrate that the innovation scatter model enables LLMs to extend innovations across structurally similar stages, thereby enhancing generalization and reuse.

\end{abstract}


\IEEEpeerreviewmaketitle

\section{Introduction}

Large Language Models (LLMs) have demonstrated impressive capabilities in generating coherent text, completing code \cite{nam2024using} \cite{su2025llm}, and performing various reasoning tasks by leveraging large-scale pretraining on diverse textual data \cite{chang2024survey}. These models operate primarily by identifying and extending statistical patterns observed in the training corpus. As a result, their success is highly dependent on the presence of analogous examples during pretraining. However, when confronted with genuinely novel input—such as a user-defined innovation not present in pretraining—LLMs often fail to generalize that innovation to structurally similar stages within a larger task or process.

In real-world applications, innovations frequently emerge in localized contexts, such as optimizations applied to a specific stage in a software pipeline, a section of a document, or a phase in a smart contract lifecycle \cite{su2024lifecycle} \cite{su2022embedding}. While such innovations are often inherently applicable to other stages due to structural or functional redundancy, current LLMs struggle to extend these changes beyond their original scope. This shortcoming arises from what we refer to as \textit{pattern dependency}—the model’s heavy reliance on previously seen templates and sequences, which constrains its ability to propagate novel patterns across unfamiliar contexts.

One key reason for this limitation is that LLMs do not perform reasoning in the human sense \cite{amirizaniani2024can}. Rather than “thinking,” an LLM relies on mappings learned from pretraining to produce likely continuations based on token prediction. This transfer mechanism captures frequent relationships between inputs and outputs but lacks the abstract reasoning required to apply new ideas systematically across multiple stages. In essence, LLMs predict rather than infer \cite{zheng2023response}. To overcome this, we propose providing the model with structured “thinking patterns” through carefully designed prompts and modular decomposition strategies, thereby enabling it to emulate higher-level reasoning behavior.

To address this issue, we propose the \textbf{innovation scatter model}—a framework designed to help LLMs generalize user-provided innovations beyond their initial scope and apply them to broader multi-stage workflows. The model decomposes the generalization process into four essential steps:

\begin{itemize}
    \item \textbf{Identification of local innovation}: The LLM is prompted to compare the new input against the context and extract the core change introduced by the user.
    \item \textbf{Generalization}: The extracted change is abstracted to eliminate references to specific stages or components, making it context-independent.
    \item \textbf{Scope determination}: The model evaluates whether the innovation applies solely to a single part or could logically extend across the overall process.
    \item \textbf{Cross-stage application}: The generalized innovation is then systematically applied to other relevant stages using guided prompts.
\end{itemize}

This approach is motivated by the observation that different stages of a structured process often share similar semantic or functional properties. Therefore, an innovation applied at one stage is likely to be beneficial elsewhere—if the model is properly guided to make that inference. By leveraging modular prompting, structural similarity, and a staged reasoning pipeline, the innovation scatter model significantly enhances the LLM’s ability to perform part-to-whole generalization.

\textbf{The contributions of this paper are summarized as follows:}

\begin{itemize}
    \item We propose a \textit{Segmental Abstraction and Transfer} (SAT) method that leverages LLMs to simulate reasoning in segmented processes—comprising either stages or parts—by abstracting innovations through the removal of segment-specific details and transferring them across segments to enable generalization. 

    \item Building on the SAT method, we introduce a structured four-step framework—referred to as the \textit{scatter model}—that systematically propagates innovations throughout multi-stage or component-based workflows. This model enables the emergence of novel solutions by removing segment-specific details, and transferring them across segments to enable generalization.

    \item We identify and formalize the \textit{pattern dependency limitation} of large language models (LLMs) in multi-stage and part-to-whole generalization tasks. We show that without structural guidance, LLMs struggle to extend localized innovations across a broader process, and we demonstrate how the scatter model addresses this challenge through explicit segmental reasoning. 
\end{itemize}

The remainder of this paper is organized as follows. Section II reviews related work. Section III introduces the scatter model. Section IV presents the verification results and corresponding analysis. Finally, Section V concludes the paper by summarizing our key contributions.

\section{Related Work}

\subsection{Pattern Dependency and Generalization in LLMs}

LLMs such as GPT-4o \cite{shahriar2024putting}, PaLM \cite{chowdhery2023palm}, and LLaMA \cite{zhu2024llama} are trained on vast corpora using next-token prediction objectives, allowing them to internalize an extensive range of syntactic and semantic patterns. These models excel at tasks involving completion, paraphrasing, and pattern continuation. However, their output behavior remains heavily constrained by the distribution of their pretraining data. This reliance has led to what is known as \textit{pattern dependency}—a phenomenon where models are only able to generate high-quality outputs when the task at hand aligns closely with patterns previously encountered during pretraining. In scenarios involving novel abstractions or out-of-distribution logic, LLMs tend to produce unstable or incoherent outputs \cite{tan2024llms}.

Several studies have investigated the limits of generalization in LLMs, particularly when asked to apply unseen transformations or synthesize novel task flows. Work by Garg et al. \cite{tan2024llms} shows that even in compositional tasks that appear trivial to humans, such as reversing syntactic structures or recombining logical components, LLMs often fail unless the patterns already exist in the training distribution. Similarly, Bommasani et al. \cite{bommasani2021opportunities} highlight that while LLMs are versatile across many benchmark tasks, they do not truly "understand" structure—they merely approximate known token sequences with high probability. This pattern-driven behavior limits their ability to generalize across stages of a workflow, particularly when user-specified innovations deviate from known formulations.

In structured environments—such as software pipelines, scientific writing, or procedural knowledge—the ability to generalize a localized innovation (e.g., a change in one section or stage) to others is critical. Yet current LLMs lack explicit mechanisms to reason about structural redundancy or infer latent similarities across stages. As a result, they may fail to apply even simple innovations consistently when those transformations have not appeared across similar contexts during pretraining. This limitation motivates the need for external guidance or frameworks—such as the proposed innovation scatter model—that can decompose, abstract, and systematically propagate changes by overcoming the inherent constraints of pattern dependency.

\subsection{Prompt Engineering and In-Context Learning}

Prompt engineering has emerged as a practical technique for steering LLM behavior without requiring parameter updates. In-context learning allows LLMs to perform new tasks by conditioning on a sequence of exemplars within the input prompt \cite{brown2020language}. This method leverages the model’s ability to recognize and replicate patterns from the prompt content, rather than relying solely on pretraining. Techniques such as zero-shot and few-shot prompting have shown promising results on a wide range of tasks, including question answering, summarization, and classification \cite{liu2023pre}. However, these approaches are highly sensitive to prompt phrasing and formatting, and typically rely on the model having seen similar tasks during training.

More advanced prompting techniques have been proposed to improve reasoning capabilities. Chain-of-thought prompting \cite{wei2022chain} encourages the model to generate intermediate reasoning steps, thereby improving accuracy on arithmetic and symbolic reasoning tasks. Self-consistency \cite{wang2023selfconsistency} further improves this by sampling multiple reasoning paths and selecting the most consistent answer. While these methods enhance the LLM's performance on complex queries, they are not specifically designed for structural generalization across stages or components. The innovation remains confined to the original input context and is not naturally transferred to related but unmentioned segments of a larger process.


\subsection{Modular Reasoning and Stage Generalization}

Modular reasoning aims to decompose complex tasks into subcomponents that can be solved independently and recombined to produce global solutions. Early work in neural module networks \cite{fashandi2023neural} demonstrated the potential of structured architectures to handle compositional queries by mapping each sub-task to a learned module. Subsequent work has extended this principle to more general domains, such as question answering and program synthesis \cite{imtiaz2023decomposing}, where reasoning steps are explicitly modeled as separate, reusable units. These approaches show that decomposability can improve generalization to unseen compositions, particularly in environments with strong structural regularities.

However, modular reasoning methods typically rely on specialized model architectures or training schemes that enforce compositional behavior, which may not be directly applicable to pre-trained autoregressive LLMs like GPT-3 or PaLM. In contrast, LLMs operate in a stateless, sequence-driven manner, lacking built-in representations of modular structure or explicit memory of previous subproblem resolutions. As a result, LLMs tend to treat each stage or part of a process independently unless guided by carefully designed prompts or external structure. Without such guidance, LLMs often replicate solutions narrowly tied to the original input, failing to propagate innovations across structurally similar parts of a task.

Recent work has attempted to bridge this gap by introducing techniques for prompting LLMs to simulate modular behavior. For example, approaches in structured editing \cite{lake2023human} encourage the model to reuse edit patterns across similar document sections or function blocks. Yet these approaches assume the existence of such patterns in pretraining data. In contrast, the innovation scatter model proposed in this paper addresses cases where the user introduces novel changes that have not previously occurred. Our method emphasizes abstraction and cross-stage transfer of innovations that arise locally but are potentially generalizable. By decoupling an innovation from its original context and explicitly applying it to other structurally aligned stages, the scatter model enables part-to-whole generalization even in the absence of pretrained support for the specific transformation.

\subsection{Contributions Beyond Existing Work}

To the best of our knowledge, this work is the first to propose a structured framework—\textit{the innovation scatter model}—for guiding LLMs to perform part-to-whole generalization of innovations via abstraction, scope analysis, and prompt-based propagation. While previous work has addressed prompting, pattern reuse, or reasoning separately, our model integrates these elements to support innovation transfer in settings where pretraining data lacks relevant examples.

\section{Scatter Model}

In many real-world applications, innovations often emerge at a single stage of a larger multi-stage process. A key question is whether such local innovations can be generalized and applied to other stages. However, Large Language Models (LLMs), which depend heavily on patterns seen during pretraining, often fail to extend new, unseen changes across different stages of a task.

This section examines the limitations of LLMs in generalizing localized innovations to broader contexts. In particular, we focus on scenarios where a novel method—unseen in the model’s pretraining data—must be extended to other parts of a structured task. These scenarios can be broadly classified into two categories: (1) those involving a \textit{temporal sequence}, where the structure is divided into distinct stages over time (with each local unit referred to as a \textbf{stage}); and (2) those involving a \textit{spatial composition}, where the structure consists of different physical or logical components (with each local unit referred to as a \textbf{part}). In both cases, the challenge lies in enabling the LLM to transfer innovations introduced in one stage or part to others that are structurally or functionally related.

For instance, in the \textit{temporal case}, if a novel optimization is applied to the deployment stage of a smart contract (such as the embedded smart contract \cite{su2022embedding}), can the same strategy be effectively extended to other stages in the contract’s lifecycle (e.g., verification, execution, or termination)? In the \textit{spatial case}, if a method is developed to improve the surface of a desk, can an LLM generalize this method to optimize other parts of the desk, such as its legs?

Let us formalize this as follows. Suppose a complex target task \( T \) can be divided into \( n \) stages:

\begin{equation} \label{eq_target_div_n1}
T = \{ t_1, t_2, ..., t_n \}
\end{equation}

If an innovation \( \Delta_i \) is introduced at stage \( t_i \), the objective is to extend this change to other relevant stages \( t_j \) (\( j \neq i \)), a process we refer to as \textit{part expansion}.

However, when such an innovation falls outside the LLM’s pretraining distribution, the model often struggles to infer and generalize the change to other parts of the task. This limitation arises from the LLM’s inherent \textbf{pattern dependency}—its reliance on known patterns seen during training—resulting in a lack of reasoning capability for previously unseen transformations. We provide a brief formal argument for this limitation below.

\subsection{Pattern Dependency and Generalization Limitations of LLMs}

LLMs generate outputs that reflect statistical patterns learned during pretraining. This section presents two foundational issues:
\begin{itemize}
    \item LLM outputs are highly dependent on pretrained patterns.
    \item LLMs struggle to generalize structural changes that deviate from those patterns.
\end{itemize}

\subsubsection{LLM Outputs Follow Pretrained Patterns}

Let \( \mathcal{D} = \{x^{(i)}\}_{i=1}^N \) denote the pretraining corpus. The model learns parameters \( \theta \) such that the next token \( x_t \) is predicted based on preceding tokens \( x_{1:t-1} \) via:

\begin{equation}
f_\theta(x_{1:t-1}) = \arg\max_{x_t} P(x_t \mid x_{1:t-1}; \theta)
\end{equation}

This optimization embeds the distributional and structural properties of the corpus \( \mathcal{D} \) into the model.

\textbf{Definition (Pattern Dependency):} Let \( \mathcal{P} \subset \mathcal{D} \) denote the set of observed patterns. Then:
\begin{equation}
\forall x \notin \text{span}(\mathcal{P}),\quad f_\theta(x) \text{ is likely unstable or incoherent}
\end{equation}

For example, if a user modifies a function name in a code block, an LLM may consistently update related references—if such co-edit patterns exist in \( \mathcal{P} \). Absent such patterns, the model may apply changes inconsistently
or not at all.

\subsubsection{Failure to Generalize Unseen Innovations Across Stages or Parts}

While LLMs can replicate and extend pretrained patterns effectively, they often fail to generalize novel changes to other parts or stages of a task when the change does not appear in the pretraining data. This limitation becomes critical in multi-stage processes, where innovations introduced in one stage may logically apply to others.

Let a task be composed of \( n \) interdependent stages or parts:
\begin{equation}
T = \{t_1, t_2, \dots, t_n\}
\end{equation}
Suppose an innovation \( \Delta_i \) is introduced at stage \( t_i \). Ideally, a transformation function \( g_j \) should propagate the effect to other stages:
\begin{equation}
\forall j \neq i,\quad \Delta_j = g_j(\Delta_i)
\end{equation}
However, if \( \Delta_i \notin \mathcal{P} \), where \( \mathcal{P} \) is the pretrained pattern space, the LLM lacks the basis to infer a coherent \( g_j \). This results in either no change or incoherent transformations across stages.

\textit{Pattern Gap.} If the LLM has never seen a pattern like \( \Delta_i \), it may attempt to approximate it by projecting onto a nearby known pattern \( p \in \mathcal{P} \):
\begin{equation}
f_\theta(\Delta_i) \approx f_\theta(p) \quad \text{where } p \in \mathcal{P},\ p \approx \Delta_i
\end{equation}
This leads to inaccurate or irrelevant outputs, especially if \( \Delta_i \) encodes domain-specific or structurally novel logic.

\textit{Structural Redundancy Across Stages.} Despite LLMs' limitations, real-world processes often feature structural redundancy across stages. That is, innovations introduced in one stage are frequently applicable to others because:
\begin{itemize}
    \item The stages represent variations of a common structure (e.g., different API phases, pipeline stages, or document sections).
    \item They share similar functional or semantic roles within the larger system.
\end{itemize}

\textbf{Definition (Innovation Redundancy Probability):} Define the probability that a stage \( t_j \) can benefit from the innovation \( \Delta_i \) as:
\begin{equation}
P(t_j \leftarrow \Delta_i) = P\big(t_j \text{ is structurally or functionally similar to } t_i \big)
\end{equation}
In structured systems, this probability is often high, even though \( \Delta_i \notin \mathcal{P} \). Thus, the LLM's inability to propagate the innovation stems not from lack of relevance, but from lack of exposure during pretraining.

Consider a smart contract lifecycle with stages \textit{deployment}, \textit{execution}, and \textit{termination}. If an innovation optimizes the deployment phase (as that of embedded smart contract \cite{su2022embedding}), a similar optimization may be applicable to execution or termination or even the whole process \cite{su2024lifecycle}. However, if such patterns do not exist in the pretraining corpus, the LLM is unlikely to make this transfer autonomously.

This illustrates that while LLMs are constrained by their pretrained patterns, the structural nature of many real-world systems suggests that innovations should naturally extend across stages. Bridging this gap requires explicit prompting, architectural mechanisms, or hybrid approaches that help the LLM recognize and apply structural analogies.

\subsubsection{Implications}

These findings reveal two core limitations:
(1) LLM outputs are fundamentally driven by pattern matching based on pretrained data.
(2) Without analogous pretrained examples, LLMs struggle to extrapolate structured innovations across multiple stages.

To address these limitations, new architectures are needed to support abstract reasoning and dynamic pattern generalization for unseen process-wide transformations.

\subsection{Applying Innovations to Entire Processes via the Scatter Model}

When a novel optimization is applied to one component or stage of a larger process, it is often desirable to propagate this innovation to other stages. For instance, consider the case of embedded smart contracts, which combine the deployment and the first invocation into a single step. Using the scatter model, we aim to generalize this optimization to other stages, such as smart contract termination, by applying similar consolidation strategies.

Let a target task \( T \) be decomposed into \( n \) stages:
\begin{equation} \label{eq_target_div_n}
  T = \{ t_1, t_2, ..., t_n \}
\end{equation}

Suppose an optimization is applied to stage \( t_i \). The goal is to extend this innovation to other stages \( t_j, t_k, \ldots \), where relevant. This process of systematically applying a local innovation across multiple stages is referred to as the \textbf{scatter model}.

\subsubsection{Scatter Model Algorithm}

The process of innovation propagation in the scatter model can be broken down into the following steps:

\begin{enumerate}[label=(\arabic*)]
    \item \textbf{Identify the Local Innovation:}
    The user provides a novel idea or optimization, denoted as \( \Delta_{\text{input}} \). The LLM is prompted to compare this with the original context \( C \), and extract the core change:
    \begin{equation}
        \Delta_{\text{local}} = \text{LLM}_{\text{diff}}(\Delta_{\text{input}}, C)
    \end{equation}
    , where \( \text{LLM}_{\text{diff}} \) represents a model behavior that performs comparison or differencing. The goal is to isolate the essential novelty introduced.

    \item \textbf{Generalize the Innovation:}
    Since \( \Delta_{\text{local}} \) may be tied to a specific part \( t_i \), this step abstracts the change by removing such dependencies. The LLM generates a generalized version:
    \begin{equation}
        \Delta_{\text{gen}} = \text{LLM}_{\text{gen}}(\Delta_{\text{local}})
    \end{equation}
    , where \( \Delta_{\text{gen}} \) is the abstracted innovation no longer bound to any specific stage or component.

    \item \textbf{Determine the Scope of Expansion:}
    Let the overall process be composed of stages \( T = \{ t_1, t_2, \dots, t_n \} \). The system evaluates whether \( \Delta_{\text{gen}} \subseteq t_i \) or \( \Delta_{\text{gen}} \subset T \). If:
    \begin{equation}
        \Delta_{\text{gen}} \subseteq t_i \quad \text{for some } i,
    \end{equation}
    then part expansion is needed. Otherwise, it may already cover multiple stages.

    \item \textbf{Execute Part Expansion:}
    For each stage \( t_j \in T, j \neq i \), apply the generalized innovation:
    \begin{equation}
        \Delta_j = \text{LLM}_{\text{apply}}(\Delta_{\text{gen}}, t_j)
    \end{equation}
    , where \( \Delta_j \) is the result of applying the generalized innovation \( \Delta_{\text{gen}} \) to another stage \( t_j \). This enables systematic propagation of the innovation across the entire process.
\end{enumerate}

Algorithm \ref{alg:scatter_expansion} outlines a four-step procedure for applying a localized innovation to multiple stages of a broader process using a Large Language Model (LLM). The method is grounded in the concept of scatter-based expansion, where an initially stage-specific change is generalized and then systematically applied across the entire structure.

\begin{algorithm}[ht]
\caption{Scatter-Based Innovation Expansion via LLM}
\label{alg:scatter_expansion}
\begin{algorithmic}[1]
\REQUIRE User input innovation \( \Delta_{\text{input}} \), original context \( C \), process stages \( T = \{t_1, t_2, \dots, t_n\} \)
\ENSURE Expanded innovations \( \{\Delta_j\} \) applied to multiple stages

\STATE \textbf{Step 1: Identify Local Innovation}
\STATE \( \Delta_{\text{local}} \leftarrow \text{LLM}_{\text{diff}}(\Delta_{\text{input}}, C) \)
\COMMENT{Extract key change from user input}

\STATE \textbf{Step 2: Generalize the Innovation}
\STATE \( \Delta_{\text{gen}} \leftarrow \text{LLM}_{\text{gen}}(\Delta_{\text{local}}) \)
\COMMENT{Remove part-specific details to obtain general form}

\STATE \textbf{Step 3: Determine Scope of Expansion}
\IF{ \( \Delta_{\text{gen}} \subseteq t_i \) for some \( t_i \in T \) }
    \STATE \textit{Mark for part expansion}
\ELSE
    \STATE \textbf{return} \( \Delta_{\text{gen}} \) \COMMENT{Already generalized}
\ENDIF

\STATE \textbf{Step 4: Execute Part Expansion}
\FOR{ each \( t_j \in T \), where \( j \neq i \) }
    \STATE \( \Delta_j \leftarrow \text{LLM}_{\text{apply}}(\Delta_{\text{gen}}, t_j) \)
\ENDFOR

\STATE \textbf{return} \( \{\Delta_j\} \)
\end{algorithmic}
\end{algorithm}


Rather than requiring the LLM to invent entirely new ideas, this approach helps it expand the scope of user-provided innovations to broader contexts, enhancing generalization.

\subsubsection{Method Generalization Strategy- Aegmental Abstraction and Transfer (SAT)}

A critical step in the scatter-based expansion process is the generalization of an innovation so that it becomes decoupled from its original context (the original stage or part). This enables the innovation to be applied beyond its initial stage or component. We define this transformation as the \textit{Segmental Abstraction and Transfer (SAT) method}. 

Let the original innovation be represented as \( \Delta_{\text{local}} \), which is inherently tied to a specific stage \( t_i \in T \). The goal of generalization is to derive a context-independent transformation \( \Delta_{\text{gen}} \) such that:

\begin{equation}
    \Delta_{\text{gen}} = \mathcal{G}(\Delta_{\text{local}}), \text{where } \mathcal{G} \text{ removes stage-specific dependencies}
\end{equation}

In practice, this is achieved by prompting the LLM to abstract away content that explicitly or implicitly refers to stage \( t_i \). This reduces the semantic coupling between the innovation and its originating stage, increasing its transferability to other stages \( t_j \in T, j \ne i \).

\textit{Definition (Coupling Strength)}. Let \( \rho(\Delta, t_i) \in [0, 1] \) denote the coupling strength between an innovation \( \Delta \) and stage \( t_i \). The generalization objective is to minimize this coupling:

\begin{equation}
    \rho(\Delta_{\text{gen}}, t_i) < \rho(\Delta_{\text{local}}, t_i)
\end{equation}

Once the coupling strength is reduced, the generalized method \( \Delta_{\text{gen}} \) can be used as input to downstream processes that apply it to other parts or stages:

\begin{equation}
    \forall t_j \in T, \quad \Delta_j = \text{LLM}_{\text{apply}}(\Delta_{\text{gen}}, t_j)
\end{equation}

\noindent This process enhances the reuse of innovations and allows structured propagation of new ideas across a process architecture.

Generalization not only supports broader applicability but also helps the LLM bypass its pattern dependency limitations by isolating transferable conceptual structures from localized instantiations.

\subsection{Component-Level Optimization}

While the scatter model is primarily designed for multi-stage processes, the same generalization framework can be extended to component-based systems where stages are not linearly ordered but structurally composed. In such systems, innovations may arise at the level of one subcomponent and hold potential applicability to others, even when those components do not participate in the same functional sequence.

Let a system \( S \) be composed of a set of interacting or coexisting components:
\begin{equation}
S = \{ c_1, c_2, \dots, c_m \}
\end{equation}
Suppose an innovation \( \Delta_i \) is introduced at component \( c_i \). The objective is to determine whether \( \Delta_i \) can be abstracted and transferred to other components \( c_j \in S, j \ne i \), based on shared properties such as material, structural symmetry, or functional role.

Consider a physical system like a desk. If a new surface-coating technique improves durability and aesthetics on the desk surface (component \( c_1 \)), that same technique may be generalized and applied to the desk legs or drawers (components \( c_2, c_3 \))—even though they were not part of the original innovation. This relies on identifying invariant features of the innovation, such as material compatibility or geometric regularity.

\textit{LLM Implications.} For LLMs, this type of component-level transfer presents challenges similar to stage-wise generalization. The model is unlikely to spontaneously apply an innovation from one subcomponent to others unless prompted to recognize shared attributes. Thus, in addition to identifying and generalizing the local innovation (as with stages), the model must be guided to map structural analogies across components:
\begin{equation}
\text{If } \phi(c_i) \approx \phi(c_j), \quad \text{then } \Delta_j \leftarrow \text{LLM}_{\text{apply}}(\Delta_i, c_j)
\end{equation}
where \( \phi(c) \) denotes the abstract structural or functional representation of component \( c \).

In software systems, this approach could allow a bug fix or performance improvement in one module (e.g., input validation) to be propagated to other modules with similar interfaces or patterns. In smart contract design, an optimization developed for an asset transfer module might also apply to voting or staking modules if they share transaction logic or lifecycle stages.

Component-level optimization generalizes the scatter model beyond linear workflows. It enables LLMs to support architectural reuse and abstraction across structurally or semantically aligned parts of complex systems. Future extensions may incorporate similarity detection metrics or embedding-based matching to automate cross-component transfer.

\section{Verification}
To evaluate the effectiveness of the proposed innovation scatter model, we conduct a verification experiment using the embedded smart contract as a representative example. Embedded smart contracts optimize a specific stage of the smart contract lifecycle—typically the deployment and first invocation phases. The full lifecycle of a smart contract interacting with the blockchain includes several key stages: deployment, instantiation, invocation (or execution), termination, and validation \cite{su2024lifecycle}. Pre-deployment stages such as code writing and compilation are not considered part of the runtime lifecycle, as they do not involve interaction with the blockchain itself.

\subsection{Experimental Setup}
We use ChatGPT (GPT-4o) as the underlying LLM for both verification strategies. Other LLMs with comparable architectural capabilities are expected to exhibit similar behavior. Two methods are evaluated and compared:

\begin{itemize}
    \item \textbf{Method A: Direct Generalization} – The LLM is prompted to generate generalized improvements based on the concept of embedded smart contracts, without any explicit guidance on stage-wise decomposition or transfer.
    
    \item \textbf{Method B: Scatter-Based Expansion} – The LLM is guided through the four-step innovation scatter model. It is first prompted to identify the key differences introduced by embedded smart contracts, and then asked whether similar optimizations can be extended to other stages of the smart contract lifecycle.
\end{itemize}

Each method is tested over 20 rounds using varied prompt phrasings to assess robustness and consistency. The outputs are evaluated using two metrics:
(1) the number of optimization items, representing distinct improvements or candidate areas for optimization identified by the LLM (i.e., how many stages or parts the LLM recognizes as potentially relevant); and
(2) the number of lifecycle stages to which these improvements are actually applied, reflecting how many distinct locations the LLM successfully transfers the innovation to.

The second metric serves as a more direct measure of generalization effectiveness, as it captures the extent to which the LLM proposes meaningful innovation propagation across other stages or components. In contrast, the first metric reflects the model's ability to recognize potential application areas, even if no explicit transfer occurs. 

\subsection{Results and Analysis}

\begin{table}[ht]
\centering
\caption{Comparison of Method A and Method B}
\label{tab:results}
\begin{tabular}{c ccc ccc}
\toprule
\textbf{Round} & \multicolumn{3}{c}{\textbf{Method A}} & \multicolumn{3}{c}{\textbf{Method B}} \\
\cmidrule(lr){2-4} \cmidrule(lr){5-7}
 & Items & Stages &  & Items & Stages &  \\
\midrule
1  & 5 & 1 & & 7 & 5 & \\
2  & 4 & 2 & & 6 & 5 & \\
3  & 5 & 2 & & 5 & 4 & \\
4  & 6 & 3 & & 6 & 4 & \\
5  & 6 & 1 & & 7 & 4 & \\
6  & 7 & 1 & & 7 & 5 & \\
7  & 5 & 2 & & 6 & 4 & \\
8  & 5 & 2 & & 6 & 4 & \\
9  & 8 & 1 & & 6 & 4 & \\
10 & 5 & 2 & & 6 & 5 & \\
11 & 6 & 2 & & 6 & 4 & \\
12 & 7 & 3 & & 6 & 5 & \\
13 & 8 & 2 & & 7 & 4 & \\
14 & 4 & 1 & & 6 & 3 & \\
15 & 5 & 3 & & 5 & 3 & \\
16 & 8 & 1 & & 6 & 4 & \\
17 & 6 & 2 & & 8 & 3 & \\
18 & 7 & 2 & & 7 & 5 & \\
19 & 5 & 1 & & 7 & 3 & \\
20 & 8 & 3 & & 6 & 4 & \\
\midrule
\textbf{Avg} & \textbf{5.85} & \textbf{1.95} & & \textbf{6.35} & \textbf{4.20} & \\
\bottomrule
\end{tabular}
\end{table}

\begin{figure*}[ht]
    \centering
    \includegraphics[width=6.5in]{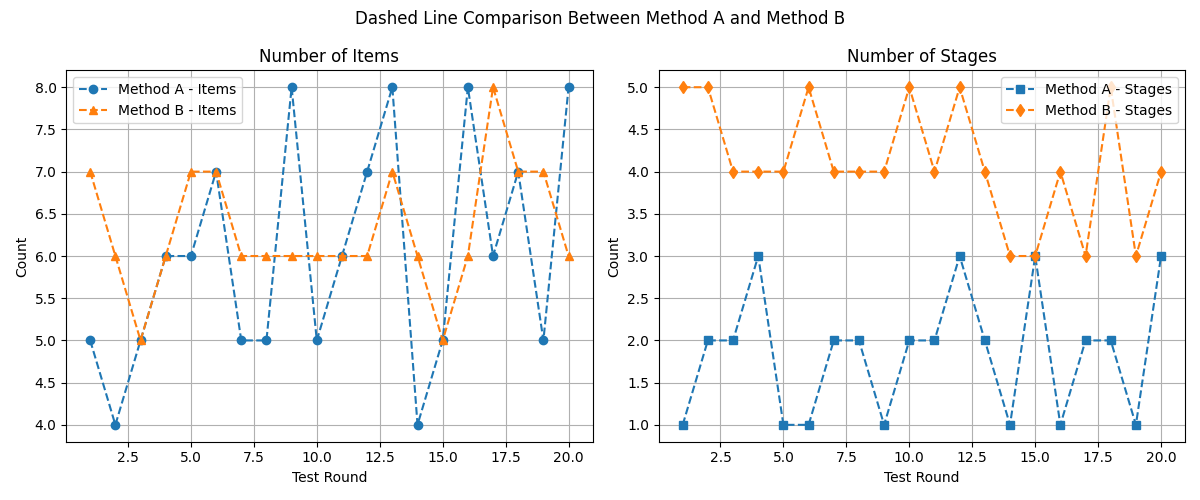}
    \caption{Comparison between Method A and Method B: Number of items and lifecycle stages considered in each round.}
    \label{exp1}
\end{figure*}

\begin{figure}[ht]
\centering
\includegraphics[width=3.4in]{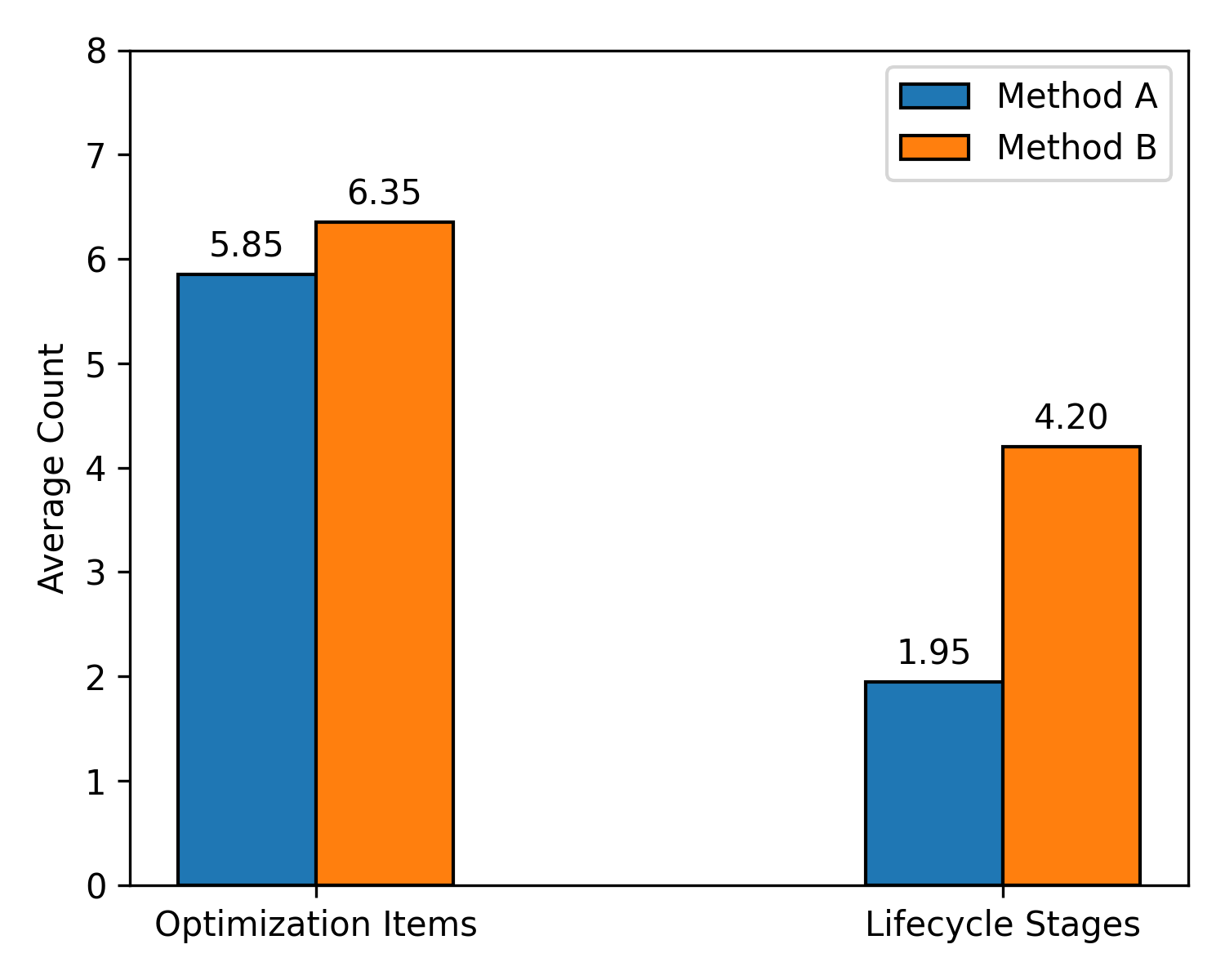}
\caption{Average number of optimization items and lifecycle stages proposed by Method A vs. Method B across 20 rounds.}
\label{fig:bar_avg}
\end{figure}

The detailed outcomes of the 20 test rounds are summarized in Table~\ref{tab:results} and visualized in Figure~\ref{exp1}. As shown, the number of optimization items proposed by Method B remains relatively stable across rounds, typically ranging from 5 to 8. This consistency suggests that the guided prompting strategy effectively helps the LLM maintain a steady level of output. In contrast, Method A exhibits greater variability (range: 4 to 8), reflecting its reliance on implicit pattern completion without structured guidance.

The most significant difference appears in \textit{stage-wise coverage}. Method A rarely applies innovations to more than three lifecycle stages, while Method B consistently extends improvements to four or five stages in most rounds. This contrast highlights the LLM’s tendency to concentrate on localized content unless explicitly directed to generalize. By guiding the model to consider broader structural parallels, Method B enables the transfer of innovations across multiple stages more effectively.

Additionally, average performance metrics reinforce these trends, as illustrated in Figure~\ref{fig:bar_avg}. Method A yields an average of 5.85 distinct optimization items and applies them to 1.95 lifecycle stages. In comparison, Method B identifies an average of 6.35 items and successfully extends them to 4.20 stages, demonstrating both higher productivity and stronger generalization capability.

These findings support the hypothesis that unguided LLMs tend to anchor their outputs within familiar, narrow contexts. In contrast, structured prompting frameworks—such as the proposed scatter model—can unlock the model’s latent reasoning capabilities and promote generalization across distinct parts of a task. This validates our core design principle: that effective cross-stage transfer in LLMs often requires explicit decomposition and guidance.

To rigorously evaluate the difference in performance between Method A and Method B, we conducted a statistical analysis on the results from the 20 test rounds. Specifically, we applied a paired two-tailed \( t \)-test to both evaluation metrics: the number of optimization items and the number of lifecycle stages covered.

Let \( X = \{x_1, x_2, \dots, x_{20}\} \) and \( Y = \{y_1, y_2, \dots, y_{20}\} \) represent the performance values of Method A and Method B respectively for each metric.

We test the following null hypotheses:
\begin{itemize}
    \item \( H_0^1 \): There is no significant difference between Method A and Method B in terms of the number of optimization items.
    \item \( H_0^2 \): There is no significant difference between Method A and Method B in terms of the number of lifecycle stages optimized.
\end{itemize}

The mean number of optimization items across all test rounds was 5.85 for Method A and 6.35 for Method B. The paired \( t \)-test yielded a p-value of \( p = 0.043 \), indicating a statistically significant improvement at the 5\% level (\( \alpha = 0.05 \)).

More notably, the mean number of lifecycle stages optimized was 1.95 for Method A and 4.20 for Method B. The corresponding \( t \)-test produced a p-value of \( p = 2.34 \times 10^{-6} \), which is highly significant (\( p < 0.001 \)), strongly rejecting the null hypothesis \( H_0^2 \).

To quantify the magnitude of improvement, we also computed Cohen’s \( d \) for effect size:
\begin{itemize}
    \item For optimization items: \( d = 0.42 \) (medium effect)
    \item For optimized stages: \( d = 1.95 \) (very large effect)
\end{itemize}

These results indicate that while Method B provides a modest improvement in the number of proposed optimizations, it achieves a substantial improvement in generalizing those optimizations across different lifecycle stages. The large effect size on the stage coverage metric supports the conclusion that scatter-based prompting enhances the generalization capabilities of LLMs in multi-stage tasks.

\section{Conclusion}

Large Language Models (LLMs) have shown remarkable performance in various natural language processing tasks, but their ability to generalize novel innovations across different stages of a structured process remains limited due to their strong dependence on pretrained patterns. This limitation becomes critical in real-world scenarios where innovations often emerge in a localized context but could logically apply across multiple stages or components.

To address this challenge, we introduced the \textit{innovation scatter model}, a four-step framework designed to guide LLMs in propagating user-defined innovations throughout a multi-stage process. By identifying the local innovation, abstracting it from stage-specific dependencies, determining its generalizability, and systematically applying it to other structurally similar parts, the scatter model enhances the reasoning and transferability capabilities of LLMs beyond their pretrained confines. Our theoretical analysis and structured prompting strategy demonstrate that part-to-whole generalization can be achieved through decomposition and guided application, even in the absence of similar patterns in training data.

Future work will explore automatic detection of transferable stages, integration with task planning modules, and broader evaluation across domains such as smart contracts, software pipelines, and technical document editing.


\ifCLASSOPTIONcaptionsoff
  \newpage
\fi

\bibliographystyle{IEEEtran}
\bibliography{ref}

%

\begin{IEEEbiography}{Hong Su}
  received the MS and PhD degrees, in 2006 and 2022, respectively, from Sichuan University, Chengdu, China. He is currently a researcher of Chengdu University of Information Technology Chengdu, China. His research interests include blockchain, cross-chain and smart contract.
\end{IEEEbiography}




\end{document}